\documentclass[pdflatex,sn-mathphys-num]{sn-jnl}% Math and Physical Sciences Numbered Reference Style
\usepackage{graphicx}
\usepackage{multirow}
\usepackage{amsmath,amssymb,amsfonts}
\usepackage{amsthm}
\usepackage{mathrsfs}
\usepackage[title]{appendix}
\usepackage{xcolor}
\usepackage{textcomp}
\usepackage{manyfoot}
\usepackage{booktabs}
\usepackage{algorithm}
\usepackage{algorithmicx}
\usepackage{algpseudocode}
\usepackage{listings}
\usepackage{dblfloatfix}
\usepackage{ulem}
\usepackage{caption}
\usepackage{subcaption}
\usepackage{dramatist}
\usepackage{xspace}
\usepackage{pifont} % http://ctan.org/pkg/pifont
\usepackage{tcolorbox}
\usepackage{xltabular}
\usepackage{longtable}
\usepackage{hyperref}
\usepackage{lineno}
\usepackage{adjustbox}
\usepackage[symbol]{footmisc}
\usepackage{setspace}
\usepackage{cleveref}
\usepackage{dsfont}
\usepackage{array}
\usepackage{tabularx}
\usepackage{lipsum}  
\usepackage{multicol}
\usepackage{titlesec}

\makeatletter
\def\abstractfont{\reset@font\fontsize{10bp}{12bp}\selectfont\leftskip=24pt\rightskip=24pt\parfillskip=0pt plus 1fil}%
\def\keywordfont{\reset@font\fontsize{10bp}{12bp}\selectfont\leftskip=24pt\rightskip=24pt plus0.5fill}%
\def\abstractheadfont{\reset@font\fontsize{12bp}{14bp}\bfseries\selectfont\titraggedcenter}%
\makeatother
\titleformat{\paragraph}[runin]{\bfseries\normalsize}{}{0pt}{}[]
\titlespacing*{\paragraph}{0pt}{0.5\baselineskip}{0.75em}
\usepackage{booktabs}   % for \toprule etc.
\usepackage{makecell}   % for multi-line headers
\usepackage{pifont}     % for checkmark and cross symbols
\usepackage{siunitx}        % For aligning numbers by decimal point
\usepackage{fontawesome5}   % Assuming you use this for the icon

\newcommand{\cmark}{\textcolor{green}{\ding{51}}} % green checkmark
\newcommand{\xmark}{\textcolor{red}{\ding{55}}}   % red cross
\theoremstyle{thmstyleone}%
\theoremstyle{thmstyletwo}%

\theoremstyle{thmstylethree}%

\begin{document}

%%=============================================================%%
%% GivenName	-> \fnm{Joergen W.}
%% Particle	-> \spfx{van der} -> surname prefix
%% FamilyName	-> \sur{Ploeg}
%% Suffix	-> \sfx{IV}
%% \author*[1,2]{\fnm{Joergen W.} \spfx{van der} \sur{Ploeg} 
%%  \sfx{IV}}\email{iauthor@gmail.com}
%%=============================================================%%

\author[1]{\fnm{Ling} \sur{Yue}}
% \email{yuel2@rpi.edu}
\equalcont{These authors contributed equally to this work.}

\author[1]{\fnm{Nithin} \sur{Somasekharan}}
% \email{somasn@rpi.edu}
\equalcont{These authors contributed equally to this work.}

\author[1]{\fnm{Tingwen} \sur{Zhang}}
% \email{zhangt20@rpi.edu}
\author[2]{\fnm{Yadi} \sur{Cao}}
% \email{yadicao95@gmail.com}
\author*[1]{\fnm{Shaowu} \sur{Pan}}\email{pans2@rpi.edu}

\affil*[1]{\orgname{\normalsize Rensselaer Polytechnic Institute}}
% \orgdiv{Department of Computer Science}, 
% \orgaddress{\street{110 8th St}, \city{Troy}, \postcode{12180}, \state{NY}, \country{USA}}

% \affil*[2]{\orgname{Rensselaer Polytechnic Institute}}
% \orgdiv{Department of Mechanical, Aerospace, and Nuclear Engineering}, 

\affil[2]{\orgname{\normalsize University of California San Diego}}
% \orgdiv{Department of Computer Science and Engineering}, 
% \orgaddress{\street{9500 Gilman Dr}, \city{La Jolla}, \postcode{92093}, \state{CA}, \country{USA}}

%%==================================%%
%% Sample for unstructured abstract %%
%%==================================%%
\newcommand{\zdaxie}[1]{\textcolor[rgb]{0.545,0,0.071}{#1}}
\newcommand{\yd}[1]{{{\textcolor{cyan}{[Yadi: #1]}}}}
\newcommand{\sw}[1]{{{\textcolor{red}{[Shaowu: #1]}}}}
\newcommand{\fa}{\textit{Foam-Agent}\xspace}
\newcommand{\fb}{\textit{Foam-Bench}\xspace}

\title{\Large Foam-Agent 2.0: An End-to-End Composable Multi-Agent Framework for Automating CFD Simulation in OpenFOAM}

\abstract{Computational Fluid Dynamics (CFD) is an essential simulation tool in engineering, yet its steep learning curve and complex manual setup create significant barriers. To address these challenges, we introduce \fa, a multi-agent framework that automates the entire end-to-end \texttt{OpenFOAM} workflow from a single natural language prompt. Our key innovations address critical gaps in existing systems: 1. An Comprehensive End-to-End Simulation Automation: \fa is the first system to manage the full simulation pipeline, including advanced pre-processing with a versatile Meshing Agent capable of handling external mesh files and generating new geometries via \texttt{Gmsh}, automatic generation of HPC submission scripts, and post-simulation visualization via \texttt{ParaView}. 2. Composable Service Architecture: Going beyond a monolithic agent, the framework uses Model Context Protocol (MCP) to expose its core functions as discrete, callable tools. This allows for flexible integration and use by other agentic systems, such as Claude-code, for more exploratory workflows. 3. High-Fidelity Configuration Generation: We achieve superior accuracy through a Hierarchical Multi-Index RAG for precise context retrieval and a dependency-aware generation process that ensures configuration consistency. Evaluated on a benchmark of 110 simulation tasks, \fa achieves an 88.2\% success rate with Claude 3.5 Sonnet, significantly outperforming existing frameworks (55.5\% for Meta\texttt{OpenFOAM}). \fa dramatically lowers the expertise barrier for CFD, demonstrating how specialized multi-agent systems can democratize complex scientific computing. The code is public at \url{https://github.com/csml-rpi/Foam-Agent}.}


%%\pacs[JEL Classification]{D8, H51}

%%\pacs[MSC Classification]{35A01, 65L10, 65L12, 65L20, 65L70}

\maketitle

\section{Introduction}\label{sec:intro}

\begin{figure}[htbp]
    \centering
    \includegraphics[width=1.0\textwidth]{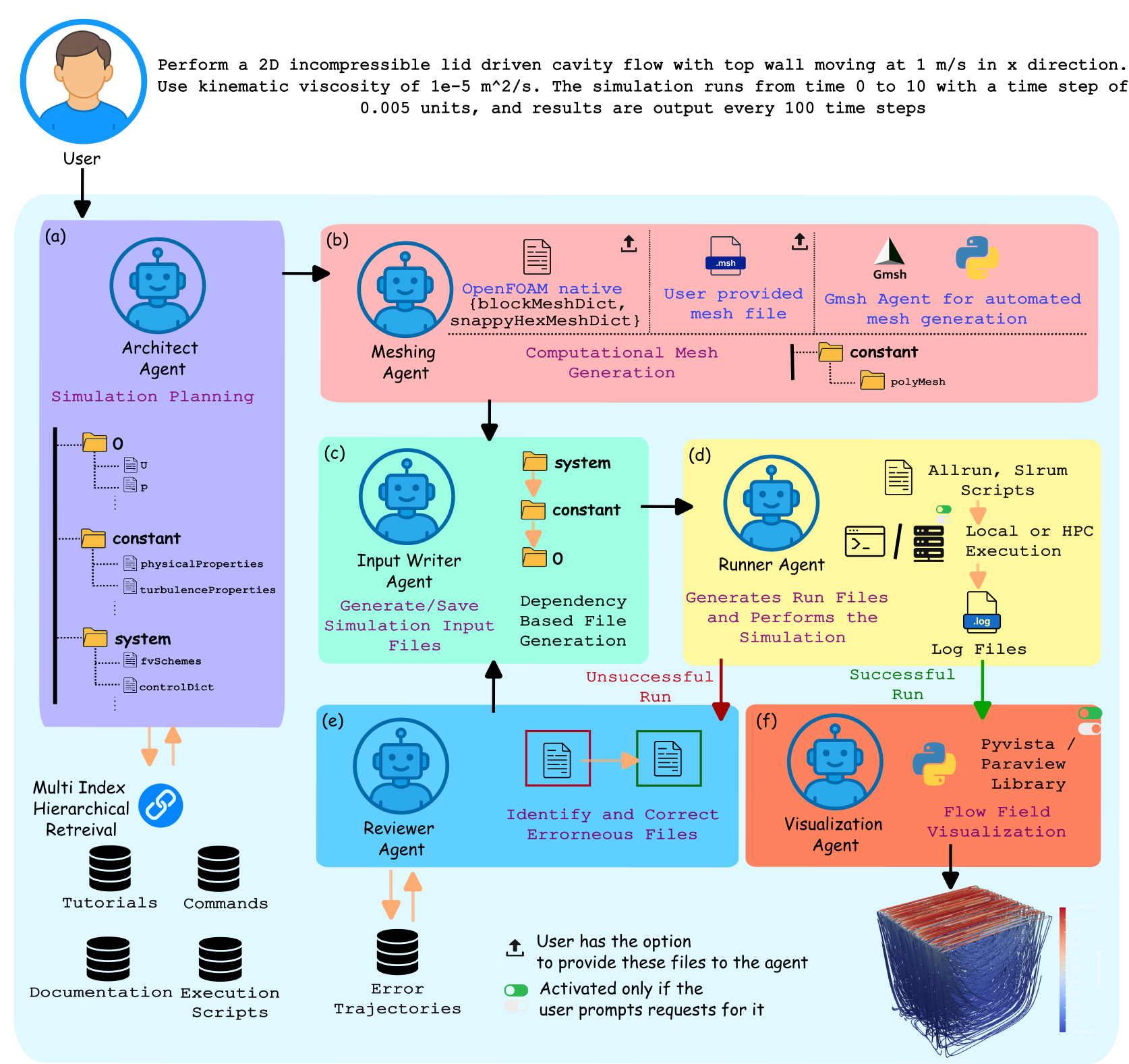}
    \caption{\fa system architecture illustrating the complete end-to-end workflow from natural language input to post-processing visualization. The system features six primary agents with dynamic adaptive workflow topology: 1. \textbf{Architect Agent}: interprets user query and plans file and folder structures to be generated, 2. \textbf{Meshing Agent}: generates the \texttt{OpenFOAM} compatible mesh required to perform the simulation. Mesh can be produced by the agent itself using \texttt{OpenFOAM} native meshing modules or by using \texttt{Gmsh} meshing library. The user also has the option to provide the agent with externally developed mesh files in the form of .msh files or OpenFOAM native \texttt{blockMeshDict}/\texttt{snappyHexMeshDict}. 3. \textbf{Input Writer Agent}: generates \texttt{OpenFOAM} configuration files required to run the simulation like \texttt{0/U, 0/p, constant/physicalProperties, system/controlDict} etc. 4. \textbf{Runner Agent} executes simulation either in the local environment of the user or in high performance computing environment as requested by the user in the prompt, 5. \textbf{Reviewer Agent}: diagnoses errors and proposes corrections through iterative debugging cycles. 6. \textbf{Visualization Agent}: generates visuals of physical quantities, if requested by the user within the user prompt, using Paraview/Pyvista python libraries. This adaptive workflow allows agents to autonomously determine execution paths based on intermediate results and problem complexity.}
    \label{fig:overview}
\end{figure}

\begin{figure}[htbp]
    \centering
    \includegraphics[width=1.0\textwidth]{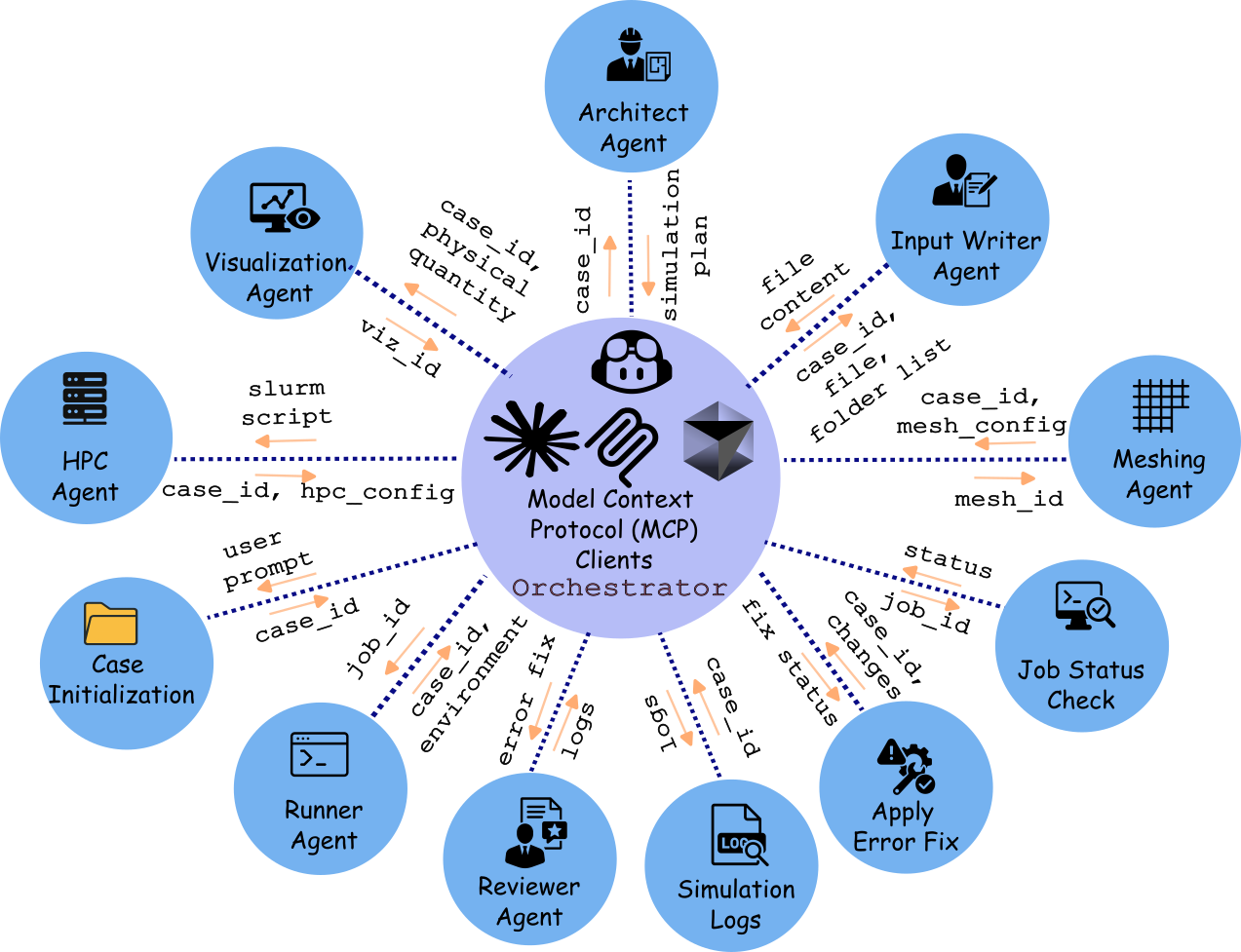}
    \caption{Modularized \fa architecture using model context protocol (MCP).}
    \label{fig:mcp}
\end{figure}

The rapid progress in generative AI, particularly large language models (LLMs), has given rise to autonomous “AI agents” capable of dramatically transforming workflows across many domains. These agents can break down complex tasks, invoke tools or other software, and iteratively refine their outputs with minimal human intervention. For example, AutoGPT \cite{Yang2023AutoGPT} allows the user to plan and execute multi-step goals (such as researching a topic or writing code) by chaining LLM queries and actions autonomously. Numerous frameworks now enable LLMs to use external tools or models like \citet{Shen2023HuggingGPT} which leverage an LLM as a central controller to orchestrate a suite of HuggingFace models for solving multi-modal tasks. Frameworks such as ReAct  \cite{yao2023reactsynergizingreasoningacting} have focused on improving the decision-making capabilities of a single agent, interleaving reasoning and acting steps, allowing an LLM to generate plans, query tools or databases, and adjust its strategy based on intermediate results. Reflexion \cite{shinn2023reflexion} introduced self-reflection and memory to enhance an agent’s ability to correct its own mistakes, while Voyager \cite{wang2023voyager} demonstrated open-ended exploration and tool use in an embodied setting like Minecraft. Toolformer \cite{schick2023toolformer} and Gorilla \cite{patil2024gorilla} extended the paradigm to reliable API calling and autonomous tool learning, ensuring agents can robustly interact with external systems. Multi-agent orchestration frameworks such as AutoGen \cite{wu2023autogen} further show how multiple LLM-based agents can collaborate to solve complex problems. Multi-agent orchestration frameworks such as AutoGen \cite{wu2023autogen} further show how multiple LLM-based agents can collaborate to solve complex problems. Beyond narrow task automation, LLM agents now demonstrate skills such as self-debugging code \cite{Dong2024SelfCode} and even autonomously conduct parts of scientific research (e.g. proposing hypotheses and analyzing data) \cite{Boiko2023Emergent}. In addition, \textit{generative agents}, which simulate believable human-like behavior, have demonstrated how AI agents can handle complex interactive workflows in virtual environments \cite{Park2023GenerativeAgents}. Together, these advances suggest that AI agents can serve as general-purpose assistants, automating or accelerating tasks that traditionally required significant human effort.

In scientific research, AI-driven agents and models have become indispensable collaborators across domains like biology, chemistry, physics, and materials science. In structural biology, AlphaFold from DeepMind \cite{Jumper2021AlphaFold} achieved atomic-level accuracy in protein structure prediction, effectively solving a long-standing challenge in the field; more recently, AlphaFold 3 extended this capability to complexes comprising proteins, nucleic acids, and ligands \cite{Abramson2024AlphaFold3}. In biomedical workflows, specialized LLM agents such as CRISPR-GPT \cite{Huang2024CRISPRGPT} automate the design of gene-editing experiments—handling tasks like guide RNA selection with minimal human input, while systems like MedAgents \cite{Tang2023MedAgents} work through diagnostic reasoning given patient information, achieving zero-shot performance on medical reasoning tasks and \citet{bionmi2025, stella2025} proposing a general-purpose biomedical agent able to carry out a spectrum of research tasks. In chemistry, systems like ChemCrow \cite{Bran2024ChemCrow} equip LLM agents with tool use to assist in synthesis planning, data querying, and reaction optimization, LLM-RDF \cite{NatureChemistry2024LLMRDF} introduces an architecture of agents tailored for experimental chemistry tasks, including design, execution, and result interpretation and \cite{elagente2025} introduces a multi-agent framework to execute quantum chemistry workflows from natural language input. In the field of materials science, several agentic frameworks have been introduced to combine available material science knowledge \cite{Hou2024HoneyComb, Buehler2024MechGPT} and provide tools for material science discovery. Further, \citet{Ghafarollahi2024Alloy} showed that a multi-agent system can be used to discover novel high-performance alloys by coordinating tasks like phase diagram reasoning and a graph neural network to evaluate candidate compounds. These categories of autonomous systems reflect the rapidly expanding footprint of AI agents across scientific workflows—transforming not just individual tasks but entire research pipelines.

The application of agentic AI extends naturally from fundamental science discovery to the application engineering domains, which is characterized by well-established but laborious workflows and a heavy reliance on domain expertise of using sophisticated specialized tools. Agents with intelligent tool calling capabilities are being developed to automate tedious and repetitive tasks such as design modifications, data collection and analysis, visualization etc., thereby scaling engineering productivity and enabling the exploration of a wider design space. Frameworks such as AutoFEA \cite{autofea2025} and MooseAgent \cite{zhang2025mooseagentllmbasedmultiagent}, have demonstrated the ability to translate natural language descriptions of engineering problems into executable input files for solvers like simulation software, CalculiX \cite{CalculiX2025},  \cite{PERMANN2020100430}, and Abaqus \cite{Smith2009_Abaqus}. These agents typically employ a step-by-step planning approach, decomposing the complex task of simulation setup into manageable sub-problems like geometry definition, meshing, material property assignment, boundary condition etc. Further, \citet{tian2024optimizingcollaborationllmbased} demonstrated the importance of optimizing the role of agents and defining clear responsibilities to each agent. The development of benchmarks like \texttt{FEABench} \cite{feabench2025} further drives progress in this area by providing a standardized means of evaluating an agent's ability to interact with FEA software APIs and solve real-world physics problems. 

Similarly, in the field of Computational Fluid Dynamics (CFD), running a simulation is still a painstaking, expertise-heavy workflow. A practitioner must define the geometry, generate a mesh, specify boundary and initial conditions, choose turbulence and physics models, and configure solver parameters. Even small errors, such as a missing field definition or inconsistent units, can cause the solver to crash, making debugging tedious and demanding deep domain knowledge. These barriers have historically limited accessibility and slowed innovation.
Recent work shows that LLM-based agentic systems are beginning to transform CFD into a conversational, automated process. Studies like \cite{chen2024metaopenfoamllmbasedmultiagentframework, pandey2025_openfoamgpt, Feng2025OpenFOAMGPT2} introduces agents, which converts natural-language problem descriptions into complete, runnable \texttt{OpenFOAM} cases by retrieving templates and embedding domain knowledge. ChatCFD \cite{Fan2025ChatCFD} offers an interactive, multi-modal conversational interface, enabling users to initialize and guide CFD simulations through. Together, these developments highlight a paradigm shift: CFD agents that reason, self-correct, and coordinate across the full pipeline, collapsing what was once days of manual setup into a single natural-language interaction.

% This broader movement towards creating agentic "copilots" for complex simulation software establishes a direct and relevant context for the automation of Computational Fluid Dynamics.

Though these studies have shown initial promise but still face major shortcomings. First, they provide incomplete workflow coverage, often focusing narrowly on solver configuration while neglecting critical pre-processing of complex geometries and meshes, as well as essential post-processing tasks like visualization. Second, they struggle with high-fidelity generation for complex cases. Finally, most are designed as single, self-contained systems, which makes them difficult to integrate with other tools or agentic systems for broader, exploratory scientific workflows.

To address these challenges and advance the emerging paradigm of AI agents with specialized tool-usage capabilities, we introduce \fa (\Cref{fig:overview}), a multi-agent framework designed to interface with OpenFOAM: a complex scientific software used in CFD. We demonstrate its ability to fully automate \texttt{OpenFOAM}-based CFD workflows while also performing the pre-processing and post-processing tasks that are essential in scientific computing. FoamAgent treats CFD simulation as a dynamic sequence of tool mediated decisions: generating and integrating computational meshes, configuring solvers, diagnosing and recovering from errors, deploying to HPC environments, and producing validated post-processed outputs. In doing so, it demonstrates how LLM based agents can not only automate solver pipelines but also write tailored inputs specific to the tool being invoked, develop auxiliary scripts for post-processing and visualization of scientific data, and orchestrate heterogeneous software components into coherent workflows. \emph{Though \fa is demonstrated on automating CFD pipelines that use \texttt{OpenFOAM}, it can still be viewed as a template for extending agentic AI frameworks to complex domains of scientific computing, where adaptability, error recovery, and end-to-end integration are essential.}

Similar to frameworks such as ChemCrow \cite{bran2024_chemcrow}, SciToolAgent \cite{ding2025_scitoolagent}, and HoneyComb \cite{Hou2024HoneyComb}, \fa leverages large language models (LLMs) as central planners that invoke external tools to accomplish complex tasks from natural language prompts, integrating specialized software into the LLM’s reasoning loop to handle domain-specific computations. It also parallels domain-focused scientific agents such as STELLA \cite{stella2025}, Biomni \cite{bionmi2025}, and AutoFEA \cite{autofea2025} in tightly coupling agentic pipelines with complex scientific ecosystems. Across these diverse systems, a common paradigm emerges: lowering expertise barriers by enabling LLMs to reason about scientific problems, retrieve relevant knowledge, and delegate subtasks to computational tools.

\fa offers several key innovations. \textbf{1. Comprehensive End-to-End Simulation Automation}: \fa is the first system to manage the full simulation pipeline, including advanced pre-processing with a versatile Meshing Agent capable of handling external mesh files and generating new geometries via \texttt{Gmsh}, automatic generation of HPC submission scripts, and post-simulation visualization via \texttt{ParaView/Pyvista \cite{paraview2005, Sullivan2019pyvista}}. \textbf{2. Composable Service Architecture}: Instead of operating as a single, self-contained agent, the framework uses the Model Context Protocol (MCP) \Cref{fig:mcp} to expose its core functions as discrete, callable tools. This design allows for flexible integration and use by other agentic systems for more exploratory workflows. \textbf{3. High-Fidelity Generation}: We achieve superior accuracy through a Hierarchical Multi-Index RAG for precise context retrieval and a dependency-aware generation process that ensures configuration consistency across all files.

Extensive experiments on a comprehensive benchmark dataset of \texttt{OpenFOAM} cases, \fa achieves an 88.2\% success rate with Claude 3.5 Sonnet, significantly outperforming existing approaches. Case studies are provided to demonstrate its ability to handle end to end workflow automation like generation/utilization of mesh files, visualization of the user specified flow fields, HPC job submission, integration of individual modules to generic agent orchestrator via MCP etc., which are the key novelties of our framework. This demonstrates the potential of specialized multi-agent systems to democratize access to complex scientific simulation tools. We public all code at \url{https://github.com/csml-rpi/Foam-Agent}.

\section{Results}\label{sec:results}

We evaluated \fa using a comprehensive benchmark dataset containing 110 \texttt{OpenFOAM} simulation cases across 11 distinct physics scenarios. The dataset spans a wide range of physical phenomena and geometric complexity \Cref{tab:physics_distribution}, providing a thorough test of automated CFD capabilities.

\begin{table}[htbp]
\centering
\caption{Distribution of benchmark cases by physics category.}
\label{tab:physics_distribution}
\begin{tabular}{lc}
\toprule
\textbf{Physics Category} & \textbf{Number of Cases} \\
\midrule
Shallow Water / Geophysical & 10 \\
Combustion / Reactive Flow  & 10 \\
Multiphase / Free Surface   & 10 \\
Shock Dynamics              & 10 \\
Turbulent Flow              & 20 \\
Laminar Flow                & 30 \\
Heat Transfer               & 20 \\
\midrule
\textbf{Total}              & \textbf{110} \\
\bottomrule
\end{tabular}
\end{table}

Each benchmark case is described using natural language prompts that include the problem description, physical scenario, geometry, solver requirements, boundary conditions, and simulation parameters. The success rate is measured by the percentage of cases that ran successfully through the agentic framework given the prompt describing the simulation scenarios. 

\paragraph{Baseline Frameworks} We compared \fa against two representative frameworks. \textbf{MetaOpenFOAM} \cite{chen2024metaopenfoamllmbasedmultiagentframework}, \textbf{OpenFOAMGPT} \cite{pandey2025openfoamgpt}. Since the authors of OpenFOAMGPT did not release their implementation, we used a variant of our own framework without the reviewer component to recreate their approach, which we refer to as \textbf{OpenFOAMGPT-Alt} throughout this paper. This implementation of our system (without the reviewer) closely mirrors their described functionality. For a comprehensive evaluation, each framework was tested with two frontier LLMs: Claude 3.5 Sonnet and GPT-4o.

% \subsubsection{Evaluation Metrics}

% We evaluated system performance using Executable Success Rate, Proportion of cases with functioning simulations.

\paragraph{Overall Performance Comparison}

\begin{table}[htbp]
\centering
\caption{Comparison of executable success rates across frameworks.}
\label{tab:framework_comparison}
\begin{tabular}{lccc}
\toprule
\textbf{Base LLM Model} & \textbf{Meta\texttt{OpenFOAM}} & \textbf{\texttt{OpenFOAM}GPT-Alt} & \textbf{\fa (Ours)} \\
\midrule
Claude 3.5 Sonnet & 55.5\% & 37.3\% & \textbf{88.2\%} \\
GPT-4o            & 17.3\% & 45.5\% & \textbf{59.1\%} \\
\bottomrule
\end{tabular}
\end{table}

\Cref{tab:framework_comparison} presents the comparative performance of the frameworks on executable success rate. \fa substantially outperforms both baselines across all tested models. With Claude 3.5 Sonnet, \fa achieves an 88.2\% success rate compared to 55.5\% for MetaOpenFOAM and 37.3\% for OpenFOAMGPT-Alt. With GPT-4o, \fa achieves 59.1\% compared to 17.3\% for MetaOpenFOAM and 45.5\% for OpenFOAMGPT-Alt.

\paragraph{Ablation Studies}\label{sec:ablation}

Having established the superiority of \fa, we delve into  the contribution of each component in our proposed framework. We analyze the impact of two key elements: the \textbf{Reviewer Node} and the \textbf{File Dependency} analysis module. All experiments were performed using the Claude-Sonnet-3.5 model. The results are summarized in \Cref{tab:ablation_study}.

\begin{table}[h!]
\centering
\caption{Ablation study results evaluating the impact of the reviewer node and file dependency analysis on success rate, token usage, and reviewer efficiency. We varied the temperature to show that \fa's performance is generalizable. The highest success rate for each configuration pair is highlighted in bold.}
\label{tab:ablation_study}
\begin{tabular}{ccc ccc}
\toprule
\makecell{\textbf{Reviewer}\\\textbf{Node}} & 
\makecell{\textbf{File}\\\textbf{Dependency}} & 
\makecell{\textbf{Temp-}\\\textbf{erature}} & 
\textbf{Success Rate (\%)} & 
\textbf{Token Usage} & 
\makecell{\textbf{Avg.}\\\textbf{Reviewer Loops}} \\
\midrule
\multicolumn{6}{c}{\textit{Configuration without Reviewer Node}} \\
\midrule
\xmark & \xmark & 0.0 & 48.2         & 282,056 & N/A \\
\xmark & \cmark & 0.0 & \textbf{56.4} & 314,670 & N/A \\
\xmark & \xmark & 0.6 & 45.4         & 282,034 & N/A \\
\xmark & \cmark & 0.6 & \textbf{57.3} & 314,922 & N/A \\
\midrule
\multicolumn{6}{c}{\textit{Configuration with Reviewer Node (max\_loops=10)}} \\
\midrule
\cmark & \xmark & 0.0 & 86.4         & 309,873 & 0.90 \\
\cmark & \cmark & 0.0 & \textbf{88.2} & 332,324 & 0.79 \\
\cmark & \xmark & 0.6 & 86.4         & 356,272 & 1.87 \\
\cmark & \cmark & 0.6 & \textbf{88.2} & 334,303 & 0.96 \\
\bottomrule
\end{tabular}
\end{table}

From the results in Table~\ref{tab:ablation_study}, it can be observed that the inclusion of the \textit{reviewer node} is the most significant factor for performance. It dramatically improves the success rate from a baseline of roughly 50\% to over 80\% across all tested configurations. This highlights the critical role of iterative feedback and self-correction in solving complex scientific computing tasks. In the absence of reviewer, \textit{file dependency} provides the most significant improvement on success rate, from 48.2\% to 56.4\% at T=0.0 and from 45.4\% to 57.3\% at T=0.6. However, because the reviewer operates independently of file dependency, the success rates do not differ significantly when the reviewer is present. As the workflow carries out more reviewer loops, any errors made during the initial file generation will be indiscriminately corrected by the reviewer. The evidence is the lower reviewer loops when file dependency is present (from 0.90 to 0.79 at T = 0.0 and from 1.87 to 0.96 at T=0.6). Therefore, the main application of file dependency is to help the reviewer converge, thus reducing API calls and workflow runtime. 

\begin{table}[h!]
\centering
\caption{Comparsion of Foam-Agent's performance using the hierarchical multi-index (hierarchy) retrieval and a single-index (baseline) retrieval. All experiments are performed using Claude-Sonnet-3.5 at T=0.6. The highest success rates are highlighted in bold.}
\label{tab:rag ablation}
\begin{tabular}{ccccc}
\toprule
\makecell{\textbf{Retrieval}\\\textbf{Method}} & 
\makecell{\textbf{Reviewer}\\\textbf{Node}} & 
\textbf{Success Rate (\%)} & 
\textbf{Token Usage} & 
\makecell{\textbf{Avg.}\\\textbf{Reviewer Loops}} \\
\midrule
\multicolumn{5}{c}{\textit{Configuration without Reviewer Node}} \\
\midrule
baseline & \xmark & 44.6 & 271,136 & N/A \\
hierarchy & \xmark & \textbf{57.3} & 306,844 & N/A \\
\midrule
\multicolumn{5}{c}{\textit{Configuration with Reviewer Node (max\_loops=10)}} \\
\midrule
baseline & \cmark & 84.6 & 323,011 & 0.73 \\
hierarchy & \cmark & \textbf{88.2} & 334,303 & 0.96 \\
\bottomrule
\end{tabular}
\end{table}
%
% https://github.com/TingwenZhang/redo-RAG/tree/main?tab=readme-ov-file#note
%

% Tingwen:

The ablation study on RAG is summarized in \Cref{tab:rag ablation}. We first performed an experiment without reviewer to isolate the effect of hierarchical multi-index retrieval. The success rate of hierarchy (57.3\%) is significantly higher than that of a single-index retrieval (44.6\%). Even with the reviewer, the effect of hierarchy is still noticible (88.2\% vs. 84.6\%). This multi-index approach significantly outperforms single-index approaches by reducing noise and improving retrieval precision. To address the semantic gap between natural language and technical terminology, we implement specialized tokenization and normalization procedures tailored to the CFD domain.

% In summary, the ablation study confirms that both the reviewer node and file dependency analysis are crucial components of our framework, with the iterative review mechanism providing the largest performance gain and the dependency analysis improving both accuracy and efficiency.

\paragraph{Case Studies} To qualitatively assess the relative accuracy of \fa and Meta\texttt{OpenFOAM}, we visually compare their simulation outputs against the Ground Truth across representative benchmark cases (\texttt{CounterFlowFlame}, \texttt{wedge} and  \texttt{forwardStep}) in \Cref{tab:framework_comparison}. The CH$_4$ mass fraction distribution at the final timestep (0.5s) in \texttt{CounterFlowFlame} is shown in the top row, the middle row depicts the temperature distribution for the \texttt{wedge} case at a timestep of 0.2 and the bottom row shows the velocity magnitude distribution for \texttt{forwardStep} case at a timestep of 0.5s. The ground truth of \texttt{CounterFlowFlame} shows a sharp, well-defined concentration gradient characteristic of flame fronts. Meta\texttt{OpenFOAM} produces a more diffuse, inaccurate transition region with considerable deviations from the expected profile. \fa's results closely match the ground truth. In the case of \texttt{wedge}, the ground truth exhibits a gradient of ve,velocity near the angled wall which is accurately captured by \fa, while Meta\texttt{OpenFOAM} fails to capture even the geometry of the problem. Finally in the case of \texttt{forwardStep}, \fa results are virtually indistinguishable from the ground truth, while Meta\texttt{OpenFOAM} results predict very low value of velocity within the domain. These cases demonstrate the accuracy of \fa in simulating canonical problems which are the stepping stones to more complicated flow fields. 

\begin{figure}[htbp]
    \centering
    \includegraphics[width=\textwidth]{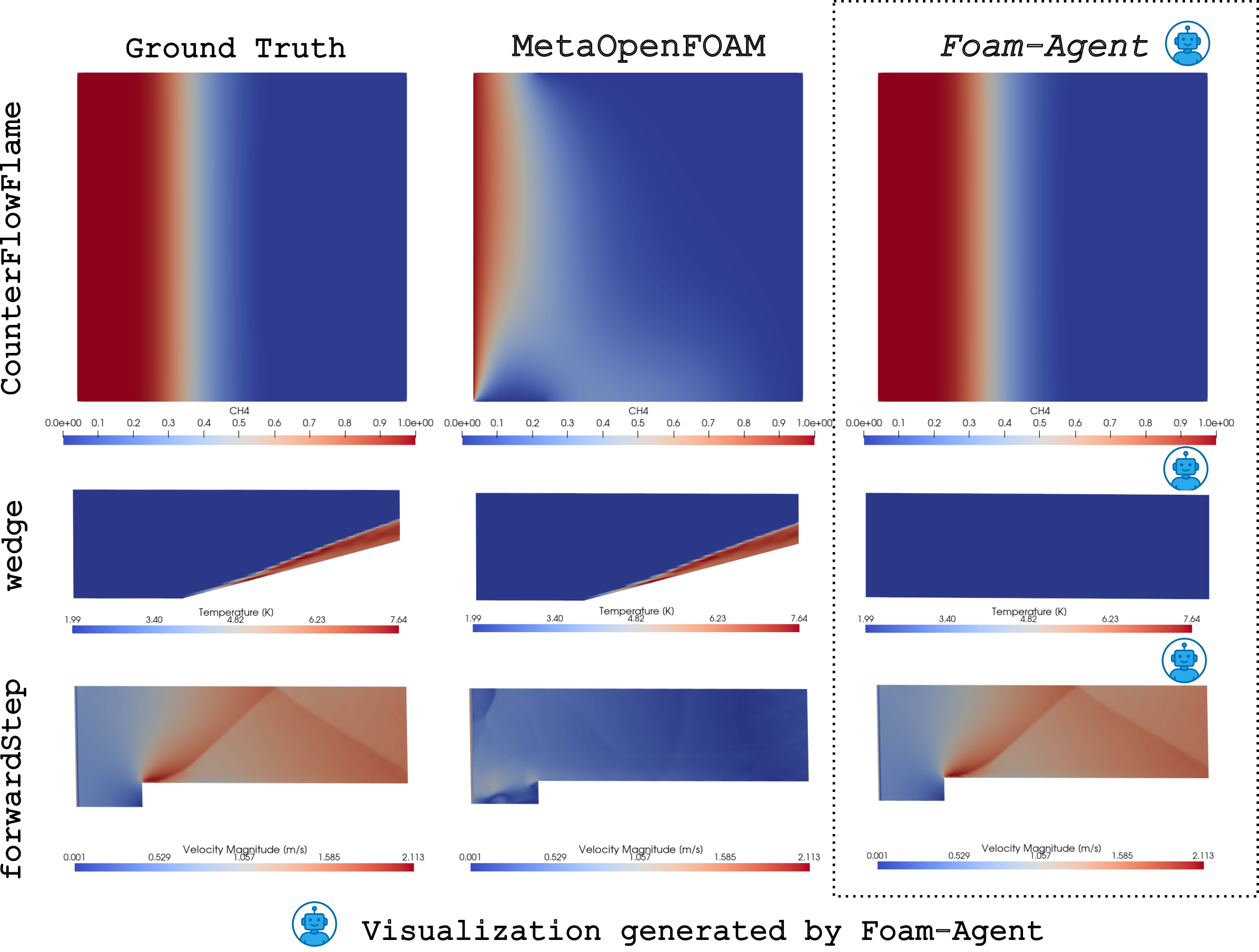}
    \caption{Comparison of simulation results produced by Meta\texttt{OpenFOAM} and \fa for \texttt{CounterFlowFlame}, \texttt{wedge} and \texttt{forwardStep} cases against the human expert generated ground truth. The top row shows the CH$_4$ mass fraction distribution comparison in \texttt{CounterFlowFlame} case at t=0.5s, the middle row shows the temperature distribution at a timestep of 0.2s for the \texttt{wedge} case and the bottom row shows the velocity magnitude distribution for \texttt{forwardStep} case at a timestep of 4s. Ground truth (left), Meta\texttt{OpenFOAM} (middle) and \fa (right). The visuals for \fa are auto generated by the agent. \fa generates a python script, which upon running loads the case and produces the visual as a \texttt{.png} file. The same python script is used to generate the visuals for ground truth and Meta\texttt{OpenFOAM} results.}
    \label{fig:framework_comaprison}
\end{figure}

% \begin{figure}[htbp]
%     \centering
%     \includegraphics[width=\textwidth]{figs/shallow_water.png}
%     \caption{Comparison of free surface height distribution in shallow water equation simulations: Ground Truth (left), Meta\texttt{OpenFOAM} (center), and \fa (right), highlighting \fa's ability to accurately reproduce complex wave dynamics.}
%     \label{fig:shallow_water}
% \end{figure}

\begin{figure}[htbp]
    \centering
    \includegraphics[width=\textwidth]{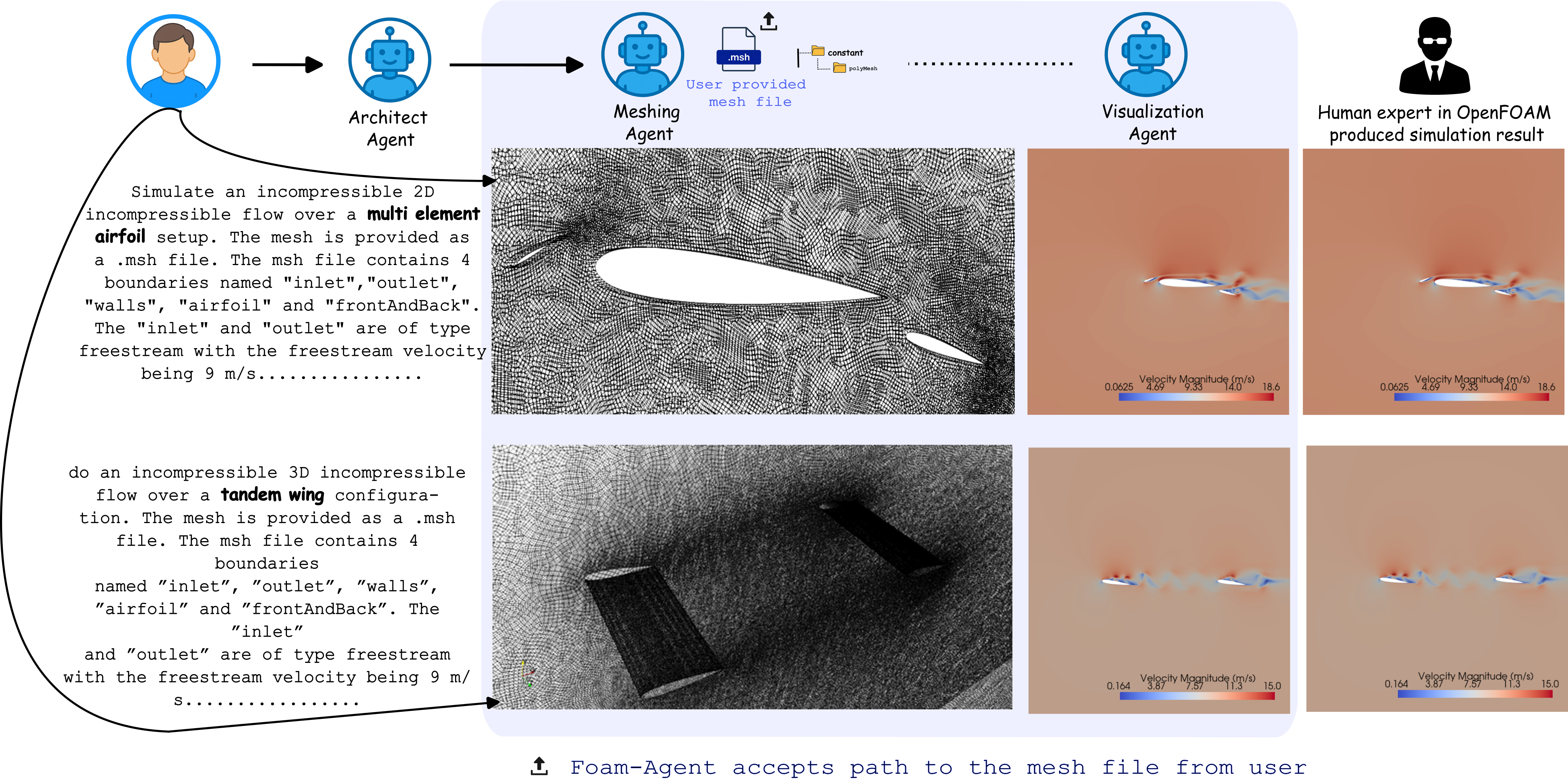}
    \caption{External mesh file processing ability of \fa analyzed through a 2D multi-element airfoil (top) and tandem wing case (bottom). The prompt describing the case along with path of the \texttt{.msh} file is given to \fa. The prompt also mentions that \fa needs to generate visuals at the final timestep for the velocity magnitude \Cref{prompt:multi_airfoil} and \Cref{prompt:tandem_wing}. The architect agent plans the required files needed for the simulation. The external mesh is converted to OpenFOAM compatible format by the meshing agent. With other agentic nodes handling the generation of input files, execution, error correction etc. At the end of workflow, the visualization agent generates a python file, which is also run by the agent to create visuals of the velocity magnitude. The same python file is used in generating the visuals of the human expert generated result to maintain consistency of the qualitative visualization.}
    \label{fig:mesh_upload}
\end{figure}

\paragraph{External Mesh File}
\textit{Ability to process externally developed mesh files is a key novelty of our framework}.  We provide mesh in the form of \texttt{.msh} files to the framework. The boundary names and flow scenario is described in the prompt to the framework. We demonstrate this functionality on two cases: 1) \textbf{Flow over multi-element airfoil} \cite{tandem_wing_repo}: This case describes the flow around multiple airfoils within the domain, with the flow of one affecting the flow of another. This 2D simulation uses an inlet velocity of 9 m/s and a fluid kinematic viscosity of $1.5\times10^{-5}$ $\mathrm{m^2/s}$. The simulation is set to use simpleFOAM solver and Spalart-Allmaras turbulence model, with a timestep of 1.0 s and a final time of 1000 s. The user prompt for the case is shown in \Cref{prompt:multi_airfoil}. 2) \textbf{Flow over tandem wing configuration} \cite{tandem_wing_repo}: This case describes the flow around a tandem configuration, with one wing located in the wake of the other. This 3D simulation uses an inlet velocity of 9 m/s and a fluid kinematic viscosity of $1.5\times10^{-5}$ $\mathrm{m^2/s}$. The simulation is set to use simpleFOAM solver and Spalart-Allmaras turbulence model, with a timestep of 1.0 s and a final time of 1000 s. The user prompt for the case is shown in \Cref{prompt:tandem_wing}. The prompts along with the path to the mesh is provided to \fa for simulating the flow field. The corresponding mesh and the contour of the velocity magnitude at the final timestep for these cases is shown in \Cref{fig:mesh_upload}. Further we also compare \fa results with human expert generated simulation results. The simulation results from \fa closely matches with the result from human \texttt{OpenFOAM} expert as shown in the velocity magnitude contour plots. Inspecting the case files generated by the agent against those prepared by humans, we found them to closely match in key aspects such as physical parameters, turbulence model selection, boundary condition assignment, and solver choice. Minor differences were observed in the numerical schemes, but these did not significantly affect the outcomes, as confirmed by the velocity contours.

% \begin{figure}[htbp]
%     \centering
%     \includegraphics[width=\textwidth]{figs/tandem_wing.png}
%     \caption{External mesh file processing ability of \fa analyzed through a 3D tandem wing case.}
%     \label{fig:viz_tandem_wing}
% \end{figure}

% \begin{figure}
%     \centering
%     \includegraphics[width=1.0\linewidth]{figs/streamlines_wing.png}
%     \caption{Streamlines over the tandem wing configuration.}
%     \label{fig:streamline_wing}
% \end{figure}

\begin{figure}
    \centering
    \includegraphics[width=1.0\linewidth]{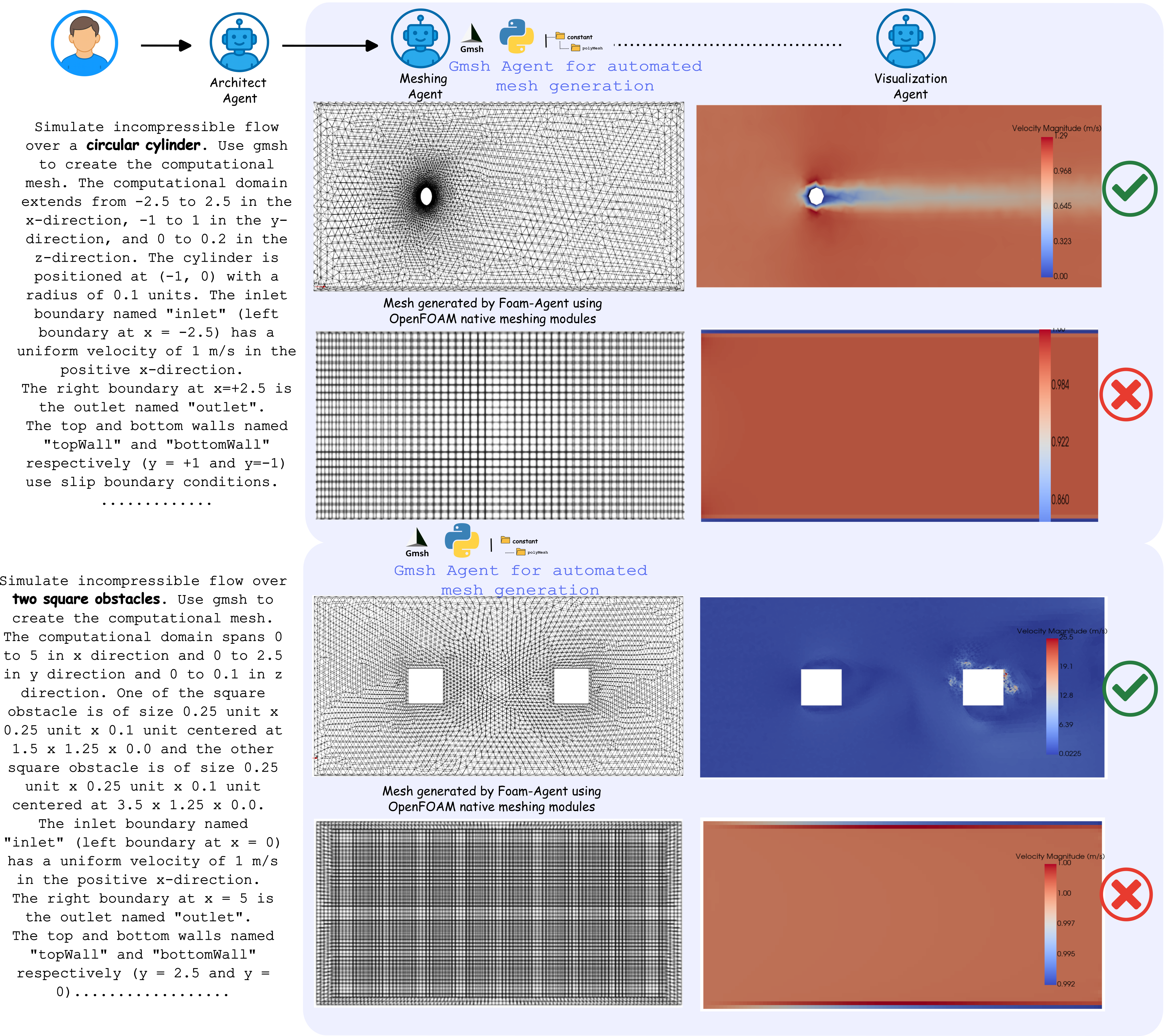}
    \caption{Flow over a cylinder (top) and flow over two square obstacle (bottom) simulated by \fa using gmsh meshing library. The user prompt describes that the mesh is to be made using \texttt{Gmsh} and the meshing agent generates the python file required to generate the mesh file utilizing the \texttt{Gmsh} library in python. The execution of the python creates the mesh in \texttt{.msh} format, which is then converted to \texttt{OpenFOAM} compatible format. The flow then proceeds to the other agents handling the generation of input files, running, error correction etc. Finally the agent creates the visuals of the velocity magnitude at the final magnitude as requested by the user within the prompt (\Cref{prompt:cylinder_flow} and \Cref{prompt:two_squares}) to \fa. To emphasize the necessity of \texttt{Gmsh} based mesh generation within the meshing agent, we show the mesh generated by \texttt{OpenFOAM's} native meshing modules given the description of the flow domain. The native meshing modules are unable to capture the obstacles in the domain and produces incorrect simulation result as shown the visualization of the velocity magnitude.} 
    \label{fig:gmsh}
\end{figure}

\paragraph{Gmsh Based Mesh Generation} We demonstrate the mesh generation capabilities of Foam-Agent utilizing Gmsh python library using two cases, where the natural language description of the mesh is provided to the agent. 1) \textbf{Flow over cylinder}: The description of the geometry given to the agent is detailed in \Cref{prompt:cylinder_flow}. 2) \textbf{Flow over two square obstacles}: The description of the geometry given to the agent is detailed in \Cref{prompt:two_squares}. The simulation is run for 10s using pisoFOAM solver. The mesh generated by the agent and the contour of velocity magnitude at the final timestep for the two cases is shown in \Cref{fig:gmsh}. To underscore the necessity of a specialized meshing agent that leverages external tools such as \texttt{Gmsh}, we compare meshes generated with \texttt{OpenFOAM}’s native meshing utilities against those produced via the Gmsh Python API. The native meshing modules were unable to capture the intended geometry of the flow scenario, whereas the Gmsh-based approach successfully generated the overall domain and accurately represented the obstacles within it. The meshing agent creates the mesh in the form of \texttt{.msh} file which is then converted to \texttt{OpenFOAM} compatible format and then further used for simulating the flow field. 

% \begin{figure}
%     \centering
%     \includegraphics[width=1.0\linewidth]{figs/two_squares.png}
%     \caption{Flow over two square onbstracles case, simulated by \fa using gmsh meshing library vs mesh generated using \texttt{OpenFOAM} native meshing files.}
%     \label{fig:two_square_case}
% \end{figure}

% \begin{figure}[h!]
%     \centering
%     \begin{subfigure}{0.45\linewidth}
%         \centering
%         \includegraphics[width=\linewidth]{figs/mesh_cylinder.png}
%         \caption{Mesh describing the cylinder geometry.}
%         \label{fig:mesh_cylinder}
%     \end{subfigure}
%     \hfill
%     \begin{subfigure}{0.45\linewidth}
%         \centering
%         \includegraphics[width=\linewidth]{figs/visualization_cylinder.png}
%         \caption{Velocity magnitude at the final timestep.}
%         \label{fig:viz_cylinder}
%     \end{subfigure}
%     \caption{Cylinder case.}
%     \label{fig:cylinder_case}
% \end{figure}

% \begin{figure}[h!]
%     \centering
%     \begin{subfigure}{0.45\linewidth}
%         \centering
%         \includegraphics[width=\linewidth]{figs/mesh_two_squares.png}
%         \caption{Mesh describing the two square obstalce geometry.}
%         \label{fig:mesh_square}
%     \end{subfigure}
%     \hfill
%     \begin{subfigure}{0.45\linewidth}
%         \centering
%         \includegraphics[width=\linewidth]{figs/visualization_two_square.png}
%         \caption{Velocity magnitude at the final timestep.}
%         \label{fig:viz_square}
%     \end{subfigure}
%     \caption{Two Squares case.}
%     \label{fig:two_squares_case}
% \end{figure}

\begin{figure}
    \centering
    \includegraphics[width=1.0\linewidth]{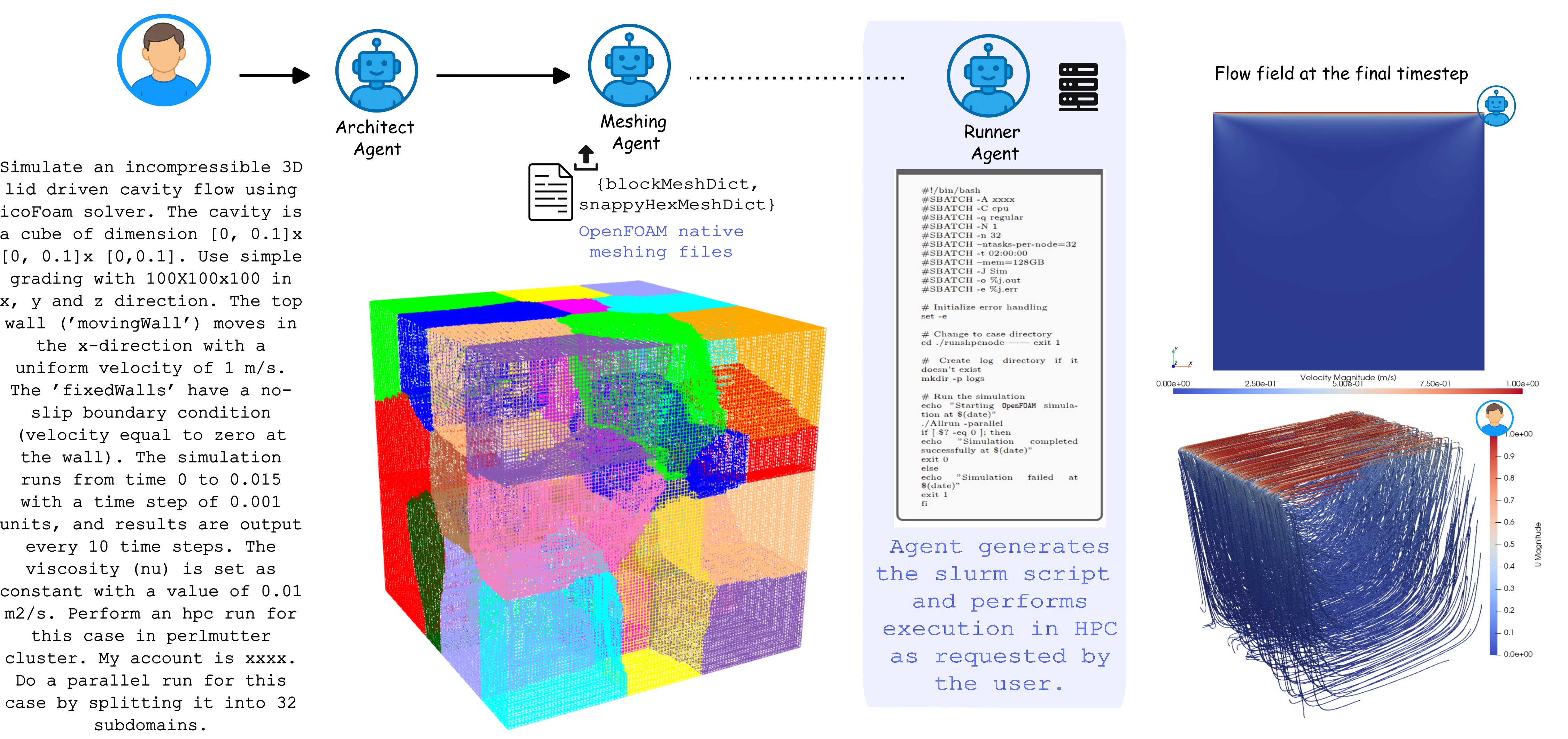}
    \caption{Highly refined simulation of a 3D lid driven cavity flow. Due to the number of cells used in the simulation the running of this case requires high performance computing (HPC) resources. The prompt suggest the cluster on which the simulation is to be run and the number of cores over which the domain is to be decomposed. The agent then creates all the required \texttt{OpenFOAM} files for the simulation along with the slurm script needed to submit the job in HPC. The mesh is divided into 32 sub-domains as mentioned in the prompt and each domain is assigned a core on which the computations for that sub-domain is carried out. Further the 2D slice of the flow field at the final timestep generated by the agent and the 3D manually generated streamlines are also shown here.}
    \label{fig:cavity_hpc}
\end{figure}

\paragraph{HPC Runner}
We demonstrate the capabilities of the HPC Runner Agent by instructing the framework to perform a 3D lid-driven cavity flow with one million cells, following the setup used in \cite{openfoam_hpc_tc2025}, using the prompt shown in \Cref{prompt:hpc_cavity}. The agent generates the necessary \texttt{OpenFOAM} case files along with a Slurm submission script for the HPC platform \textit{Perlmutter}. The platform name and submission account number are provided in the prompt, while the agent relies on the LLM’s knowledge of the system’s documentation to produce correct Slurm syntax, which can vary across clusters. The resulting highly refined mesh, the velocity contours at the final timestep (mid–$z$ slice) and corresponding streamlines are presented in \Cref{fig:cavity_hpc}. The automatically generated Slurm script is given in \Cref{slurm_script}.

\section{Methods}\label{sec:method}

\subsection{System Architecture Overview}

\fa employs a modular architecture comprising six primary components. This multi-agent framework interprets the natural language requirement from the user, leverages Retrieval-Augmented Generation (RAG) to find similar cases and configurations within tutorial database, generates simulation configurations, execute simulations, analyze errors, and implement corrections through iterative refinement. The six primary components form the backbone of the system are \textsc{Architect Agent}, \textsc{Meshing Agent}, \textsc{Input Writer Agent}, \textsc{Runner Agent}, \textsc{Reviewer Agent}, \textsc{Visualization Agent}.

\subsection{Agent Components}

\paragraph{Architect Agent} The Architect Agent translates natural language descriptions into structured simulation plans through a three-stage process. First, in the Requirement Classification stage, it analyzes the user requirement to classify the simulation according to domain-specific taxonomies, employing Pydantic models for structured output validation. Second, during Reference Case Retrieval, it queries the hierarchical indices to identify semantically similar cases, using a cascading approach that refines initial matches with detailed structural information. Finally, in the Simulation Decomposition stage, it decomposes the task into required files and directories, creating a detailed plan specifying file dependencies and generation priorities. The output for a simulation requirement $R$ is a structured plan $P = \{F_1, F_2, ..., F_n\}$ where each $F_i$ represents a required file with its dependencies, content requirements, and generation priority.

\paragraph{Meshing Agent} There multiple ways that \fa handles the meshing (pre-processing) stage of the simulation: \textbf{1) \texttt{OpenFOAM} native:} The agent generated the required \texttt{OpenFOAM} native files for mesh generation like \texttt{blockMeshDict} and/or \texttt{snappyHexMeshDict}. It then continues to generate the polyMesh folder which contains the detailed mesh information for \texttt{OpenFOAM} to process the case by running commands such as \texttt{blockMesh} and/or \texttt{snappyHexMesh}. The agent has the complete autonomy in this scenario. \textbf{2) External Mesh Files:}  
\fa does not generate the mesh files for the simualtion. The user has the ability to provide the agent with mesh specific files either in the form of native \texttt{OpenFOAM} dictionaries (\texttt{blockMeshDict}, \texttt{snappyHexMeshDict}) or pre-generated meshes from external tools (e.g., \texttt{.msh}). Given any of these inputs, the agent converts the mesh to \texttt{OpenFOAM} format, generating the ployMesh folder. \textbf{3) Gmsh:} \fa can take in natural language input of the physical domain, the boundary names and generate a mesh file (.msh) based on this description. The agent creates a python script representing the geometry and mesh using Gmsh python library and with further execution generating the mesh file. This mesh file is then converted to \texttt{OpenFOAM} format. \textit{This functionality was found be to required as \texttt{OpenFOAM}'s inherent meshing creation capabilities were found to be insufficient in creating novel geometries not present in the tutorials.} The agent intelligently chooses one of these options based on the user requirements. 

\paragraph{Input Writer Agent} Following established human CFD expertise, the Input Writer Agent implements a structured file generation sequence that respects \texttt{OpenFOAM}’s hierarchical organization: it begins with the \texttt{system} directory (simulation control parameters and numerical schemes), proceeds to the \texttt{constant} directory (physical, turbulence properties), then the \texttt{0} directory (initial and boundary conditions), and finally produces auxiliary files (e.g., \texttt{Allrun}) for executing the simulation. This ordering naturally enforces dependencies, as files prescribing boundary conditions depend on turbulence model/physical properties provided in the \texttt{constant} directory, which themselves can be dependent on solver configurations. Formally, the process can be represented as a directed acyclic graph $G = (V, E)$, where vertices $V$ denote files and edges $E$ capture their dependencies. To maintain internal consistency, the agent employs \textit{Contextual Generation}, where the context $C_i$ of each file $F_i$ includes all previously generated files, enabling coherent parameter referencing across the configuration. Finally, schema validation is applied through Pydantic models, ensuring that generated files satisfy both the syntactic and semantic constraints of \texttt{OpenFOAM}, thereby reducing generation errors.

\paragraph{Runner Agent} The Runner Agent interfaces with the \texttt{OpenFOAM} execution environment by preparing the simulation (cleaning artifacts, setting up output capture) and running the \texttt{Allrun} script. Simulations can be executed either on the user’s \textit{local machine} or deployed to \textit{HPC clusters}, as specified in the requirements. For HPC runs, the agent automatically generates Slurm scripts, submits jobs with the provided account number, and monitors their progress until completion. It can use prompt-specified parameters (e.g., nodes, processes per node) or infer them from the mesh decomposition and problem size. This capability enables \fa to scale seamlessly from desktop prototyping to large-scale industrial CFD workloads. At the end of simulation, it performs critical error detection by analyzing these logs to identify specific error patterns, extracting relevant messages and contextual information for subsequent analysis by other components. The error detection process is formalized as a pattern matching function $E: L \rightarrow \{e_1, e_2, ..., e_m\}$ that maps execution logs $L$ to a set of structured error records $e_i$, each containing the error message, location, and severity.

\paragraph{Reviewer Agent} The Reviewer Agent implements error analysis and correction of the case being run. It performs \textit{Error Contextualization} by assembling detailed context including error messages, affected files, and reference cases. It then conducts \textit{Review Trajectory Analysis} by maintaining a record of previous attempts as $H = \{(F_i^1, E^1), (F_i^2, E^2), \ldots, (F_i^n, E^n)\}$, where $F_i^j$ represents the file state in iteration $j$ and $E^j$ represents the resulting errors. 
Finally, it executes \textit{Solution Generation} by producing structured modifications to problematic files while maintaining consistency with other configuration elements. The correction process can formalized as an optimization problem: finding the minimal set of changes $\Delta F$ to files $F$ that eliminates errors $E$ while respecting user constraints $C$ and maintaining consistency across all files (\Cref{alg:iterative_refinement}). This review–correction cycle is repeated iteratively until either the errors are resolved or a maximum number of attempts specified by the user is reached (\Cref{fig:overview}).

\paragraph{Visualization Agent} After the runner agent produces a successful run, \fa verifies if visualization is requested by the user within the provided prompt. If yes, the agent will try to understand the quantity to be visualized from the prompt and generate a python script utilizing either Pyvista library or paraview python routines (the user can choose) to generate the required visualization. The agent will then execute the python file and correct errors if any (like the reviewer agent) till the visualization(s) is saved as a .png file in the run directory. The maximum number of tries can be set by the user.

\begin{algorithm}
\caption{Iterative Refinement Process}
\label{alg:iterative_refinement}
\begin{algorithmic}[1]
\Require Natural language requirement $R$, maximum iterations $M$
\Ensure Final simulation configuration $F^*$
\State $P \gets \operatorname{Architect}(R)$ \Comment{Generate initial plan}
\State $G \gets \operatorname{Meshing}(R)$ \Comment{Generate Computational Mesh}
\State $F^0 \gets \operatorname{InputWriter}(P)$ \Comment{Initial file generation}
\State $H \gets \{\}$ \Comment{Initialize history}
\For{$i \gets 1$ to $M$}
    \State $L^i, S^i \gets \operatorname{Runner}(F^{i-1})$ \Comment{Execute simulation}
    \If{$S^i = \text{SUCCESS}$}
        \State \Return $F^{i-1}$ \Comment{Simulation successful}
    \EndIf
    \State $H \gets H \cup \{(F^{i-1}, L^i)\}$ \Comment{Update history}
    \State $E^i \gets \operatorname{ParseErrors}(L^i)$ \Comment{Extract errors}
    \State $\Delta F^i \gets \operatorname{Reviewer}(E^i, F^{i-1}, H)$ \Comment{Generate corrections}
    \State $F^i \gets \operatorname{Apply}(F^{i-1}, \Delta F^i)$ \Comment{Apply corrections}
\EndFor
\State \Return $F^M$ \Comment{Return best attempt}
\end{algorithmic}
\end{algorithm}

\subsection{RAG}
A key innovation in \fa is its hierarchical multi-index retrieval system (\Cref{algo:hmir}) that segments domain knowledge into specialized indices optimized for specific phases of the simulation workflow. This approach significantly improves retrieval precision compared to conventional single-index RAG systems.

\paragraph{Knowledge Base Organization} We construct the knowledge base by parsing \texttt{OpenFOAM}'s tutorial cases, extracting information across four dimensions. The first dimension is \texttt{Case Metadata}, which includes fundamental attributes such as case name, flow domain, physical category, and solver selection. The second dimension is \texttt{Directory Structures}, which captures the hierarchical organization of files and directories in reference cases. The third dimension is \texttt{File Contents}, which preserves configuration file content, including syntax, parameter definitions, and commenting. The fourth dimension is \texttt{Execution Scripts}, which includes command sequences for preparation, execution, and post-processing.

\paragraph{Specialized Vector Index Architecture} Rather than using a monolithic database, \fa implements four distinct FAISS \cite{douze2024faiss} indices, each serving a specific purpose. The \texttt{Tutorial Structure Index} encodes high-level case organization patterns for identifying appropriate structural templates. The \texttt{Tutorial Details Index} contains configuration details for boundary conditions, numerical schemes, and physical models. The \texttt{Execution Scripts Index} stores execution workflows for generating appropriate command sequences. The \texttt{Command Documentation Index} maintains utility documentation for correct command usage and parameter selection. Each index employs a 1536-dimensional vector with \texttt{text-embedding-3-small} model from OpenAI. The retrieval process is given in \Cref{algo:hmir}.

\begin{algorithm}
\caption{Hierarchical Multi-Index Retrieval}
\label{algo:hmir}
\begin{algorithmic}[1]
\Require Query $q$, stage $s$, previous context $c$
\Ensure Retrieved context $r$
\State $E \gets \operatorname{Embed}(q)$ \Comment{Generate query embedding}
\State $I \gets \operatorname{SelectIndex}(s)$ \Comment{Select appropriate index for stage}
\State $R_i \gets \operatorname{TopK}(I, E, k=5)$ \Comment{Retrieve top-k matches}
\State $R_f \gets \operatorname{FilterByRelevance}(R_i, q, c)$ \Comment{Filter by relevance}
\State $r \gets \operatorname{FormatContext}(R_f, s)$ \Comment{Format based on stage}
\State \Return $r$
\end{algorithmic}
\end{algorithm}

\subsection{Decoupling Capabilities via a Model Context Protocol (MCP)}

To transition \fa from a monolithic tool into a flexible scientific service, we design its core around the Model Context Protocol (MCP). This decouples the CFD workflow into atomic, callable functions exposed via a standardized protocol, making \fa a composable component that higher-level agents or workflow engines can orchestrate. The MCP design follows three principles: \textbf{atomicity} (each function does one task), \textbf{statefulness} (tracking multi-stage simulations via identifiers such as case\_id and job\_id etc.), and \textbf{workflow decoupling} (separating meshing, solving, and post-processing). These features maximize flexibility while preserving fine-grained control, with the key functions summarized in \autoref{tab:mcp_functions}.

\begin{table*}[!htbp]
\small
\centering
\caption{The core functions of the \fa Model Context Protocol (MCP). Each function represents a decoupled capability within the CFD workflow, featuring strongly-typed inputs and outputs to ensure reliable interaction with orchestrating agents.}
\label{tab:mcp_functions}
\resizebox{\textwidth}{!}{%
\begin{tabular}{>{\raggedright\arraybackslash}p{0.3\textwidth} >{\raggedright\arraybackslash}p{0.35\textwidth} >{\raggedright\arraybackslash}p{0.15\textwidth} >{\raggedright\arraybackslash}p{0.15\textwidth}}
\toprule
\textbf{Function Name} & \textbf{Description} & \textbf{Input Schema} & \textbf{Output Schema} \\
\midrule
\texttt{create\_case} & Initializes a new CFD simulation case and its workspace. & \texttt{\{user\_prompt: str\}} & \texttt{\{case\_id: str\}} \\
\addlinespace
\texttt{plan\_simulation\_structure} & (Architect Agent) Plans the required file and directory structure based on the user prompt. & \texttt{\{case\_id: str\}} & \texttt{\{plan: List[\{file, folder\}]\}} \\
\addlinespace
\texttt{generate\_file\_content} & (Input Writer Agent) Generates the content for a single specified configuration file. & \texttt{\{case\_id, file, folder\}} & \texttt{\{content: str\}} \\
\addlinespace
\texttt{generate\_mesh} & (Meshing Agent) Asynchronously generates the computational mesh using a specified method. & \texttt{\{case\_id, mesh\_config: Dict\}} & \texttt{\{job\_id: str\}} \\
\addlinespace
\texttt{generate\_hpc\_script} & (HPC Agent) Generates a job submission script (e.g., Slurm) for a high-performance computing cluster. & \texttt{\{case\_id, hpc\_config: Dict\}} & \texttt{\{script\_content: str\}} \\
\addlinespace
\texttt{run\_simulation} & (Runner Agent) Asynchronously executes the simulation either locally or by submitting to an HPC cluster. & \texttt{\{case\_id, environment: str\}} & \texttt{\{job\_id: str\}} \\
\addlinespace
\texttt{check\_job\_status} & Checks the status of any asynchronous job (meshing, simulation, visualization). & \texttt{\{job\_id: str\}} & \texttt{\{status: Dict\}} \\
\addlinespace
\texttt{get\_simulation\_logs} & Retrieves detailed logs for a failed job to enable error diagnosis. & \texttt{\{case\_id, job\_id\}} & \texttt{\{logs: Dict\}} \\
\addlinespace
\texttt{review\_and\_suggest\_fix} & (Reviewer Agent) Analyzes error logs and proposes corrective actions. & \texttt{\{case\_id, logs\}} & \texttt{\{suggestions: Dict\}} \\
\addlinespace
\texttt{apply\_fix} & Applies suggested modifications to the relevant case files. & \texttt{\{case\_id, modifications: List\}} & \texttt{\{status: str\}} \\
\addlinespace
\texttt{generate\_visualization} & (Visualization Agent) Asynchronously generates a visualization of the simulation results. & \texttt{\{case\_id, quantity, ...\}} & \texttt{\{job\_id: str\}} \\
\bottomrule
\end{tabular}%
}
\end{table*}

While the MCP provides the foundational capabilities, a robust framework is required to orchestrate these functions. We achieve this through the following.

\paragraph{Stateful Workflow Orchestration with LangGraph} The sequence of MCP function calls is not fixed; it is dynamically determined by an intelligent orchestrator. We implement this orchestrator as a stateful graph using LangGraph. The nodes in the graph correspond to calls to the MCP functions, while the edges represent conditional logic that directs the workflow based on the outcomes of each step.

% This architecture is particularly effective for implementing the system's core iterative refinement loop. For instance, upon a simulation failure detected via \texttt{check\_job\_status}, a conditional edge routes the workflow to the debugging subgraph (\texttt{get\_simulation\_logs} $\rightarrow$ \texttt{review\_and\_suggest\_fix} $\rightarrow$ \texttt{apply\_fix}) before attempting to run the simulation again. 

\paragraph{Ensuring Reliability with Structured I/O}
A key challenge in LLM–tool interaction is avoiding errors from malformed or inconsistent data. To ensure reliability, we enforce strict schemas for all data exchanges, including MCP function I/O and LangGraph state. Using Pydantic, we define explicit data models that enable runtime validation and type checking, establishing a clear contract between the LLM, orchestrator, and tool functions.
% By catching data-related errors early and ensuring all components communicate via a validated, structured format, we significantly enhance the robustness and predictability of the entire framework.

\paragraph{Achieving Observability with LangSmith}
The complex and often non-deterministic behavior of multi-agent systems makes them notoriously difficult to debug and analyze. To address this, we integrate LangSmith for end-to-end traceability. Every MCP function call, LLM invocation, and state transition within the LangGraph orchestrator is logged into an immutable record of the agent's "thought process" and actions. This allows precise reconstruction of a simulation’s execution, including intermediate inputs, outputs, and errors, providing the observability needed for debugging, performance analysis, and ensuring scientific reliability.

% \section{Discussion}\label{sec:disc}

% Our evaluation of \fa demonstrates remarkable advancements in physics-based simulations. The framework achieves an impressive 88.2\% success rate, substantially outperforming the best baseline system at 55.5\% across various LLMs and physics categories. Ablation analysis confirms that each component contributes meaningfully to overall performance, with the error correction mechanism proving most critical—its removal alone causes a 55.4\% performance drop. Despite these achievements, \fa still faces challenges with complex physical phenomena involving chemical reactions or multi-phase interactions where specialized knowledge remains incomplete. Additionally, while handling standard geometries effectively, the system encounters difficulties with novel or highly complex geometrical configurations during mesh generation. The computational cost of the iterative refinement process also presents practical limitations for time-sensitive applications, as complex simulations may require multiple resource-intensive iterations.

\section{Conclusion}\label{sec:conc}

This work introduced \fa, a modular multi-agent framework that automates end-to-end CFD workflows using \texttt{OpenFOAM} from natural language prompts. Our benchmark evaluation across 110 diverse cases demonstrated an 88.2\% success rate, substantially surpassing prior frameworks. Extending beyond its core capabilities, \fa introduces integrated pre-processing, allowing users to import external mesh files or generate meshes on demand using external tools such as \texttt{Gmsh}, while also providing seamless execution on high-performance computing (HPC) platforms. Its decoupled architecture, built on the Model Context Protocol and LangGraph orchestration, enables seamless integration with other agentic ecosystems. All these capabilities not only lower the barrier of entry for non-experts but also empower experienced engineers to accelerate complex workflows. \fa establishes a foundation for scalable and trustworthy scientific automation. Future directions include extending its modular services to other solvers, enhancing physics coverage. Ultimately, this framework represents a step toward democratizing access to high-fidelity simulation and shaping the role of intelligent agents in computational science.

\backmatter

% \bmhead{Supplementary information}

% If your article has accompanying supplementary file/s please state so here. 

% Authors reporting data from electrophoretic gels and blots should supply the full unprocessed scans for key as part of their Supplementary information. This may be requested by the editorial team/s if it is missing.

% Please refer to Journal-level guidance for any specific requirements.

\bmhead{Acknowledgements}

This work was supported by U.S. Department of Energy under Advancements in Artificial Intelligence for Science with award ID DE-SC0025425. 
The authors thank the Center for Computational Innovations (CCI) at Rensselaer Polytechnic Institute (RPI) for providing computational resources during the early stages of this research. Numerical experiments are performed using computational resources granted by NSF-ACCESS for the project PHY240112 and that of the National Energy Research Scientific Computing Center, a DOE Office of Science User Facility using the NERSC award NERSC DDR-ERCAP0030714.
S. Pan is supported by Google Research Scholar Program. 

% \section*{Declarations}

% Some journals require declarations to be submitted in a standardised format. Please check the Instructions for Authors of the journal to which you are submitting to see if you need to complete this section. If yes, your manuscript must contain the following sections under the heading `Declarations':

% \begin{itemize}
% \item Funding
% \item Conflict of interest/Competing interests (check journal-specific guidelines for which heading to use)
% \item Ethics approval and consent to participate
% \item Consent for publication
% \item Data availability 
% \item Materials availability
% \item Code availability 
% \item Author contribution
% \end{itemize}

% \noindent
% If any of the sections are not relevant to your manuscript, please include the heading and write `Not applicable' for that section. 

%%===================================================%%
%% For presentation purpose, we have included        %%
%% \bigskip command. Please ignore this.             %%
%%===================================================%%
% \bigskip
% \begin{flushleft}%
% Editorial Policies for:

% \bigskip\noindent
% Springer journals and proceedings: \url{https://www.springer.com/gp/editorial-policies}

% \bigskip\noindent
% Nature Portfolio journals: \url{https://www.nature.com/nature-research/editorial-policies}

% \bigskip\noindent
% \textit{Scientific Reports}: \url{https://www.nature.com/srep/journal-policies/editorial-policies}

% \bigskip\noindent
% BMC journals: \url{https://www.biomedcentral.com/getpublished/editorial-policies}
% \end{flushleft}

\clearpage

\bibliography{sn-bibliography}% common bib file

\clearpage

\begin{appendices}

\section{System and User Prompts}

This appendix presents all system and user prompts used in the Foam-Agent framework for various components. These prompts are organized by agent role and function within the multi-agent architecture.

\subsection{Architect Agent Prompts}

The Architect Agent interprets user requirements into a structured simulation plan and breaks down complex tasks into manageable subtasks.

\subsubsection{Case Description Prompts}

\begin{tcolorbox}[colback=blue!5!white,colframe=blue!75!black,title=Case Description System Prompt]
Please transform the following user requirement into a standard case description using a structured format.
The key elements should include case name, case domain, case category, and case solver.
Note: case domain must be one of [case\_stats['case\_domain']].
Note: case category must be one of [case\_stats['case\_category']].
Note: case solver must be one of [case\_stats['case\_solver']].
\end{tcolorbox}

\begin{tcolorbox}[colback=blue!5!white,colframe=blue!75!black,title=Case Description User Prompt]
User requirement: \{user\_requirement\}.
\end{tcolorbox}

\subsubsection{Task Decomposition Prompts}

\begin{tcolorbox}[colback=blue!5!white,colframe=blue!75!black,title=Task Decomposition System Prompt]
You are an experienced Planner specializing in \texttt{OpenFOAM} projects. 
Your task is to break down the following user requirement into a series of smaller, manageable subtasks. 
For each subtask, identify the file name of the \texttt{OpenFOAM} input file (foamfile) and the corresponding folder name where it should be stored. 
Your final output must strictly follow the JSON schema below and include no additional keys or information:

\{
  "subtasks": [
    \{
      "file\_name": "$<$string$>$",
      "folder\_name": "$<$string$>$"
    \}
    // ... more subtasks
  ]
\}

Make sure that your output is valid JSON and strictly adheres to the provided schema.
Make sure you generate all the necessary files for the user's requirements.
\end{tcolorbox}

\begin{tcolorbox}[colback=blue!5!white,colframe=blue!75!black,title=Task Decomposition User Prompt]
User Requirement: \{user\_requirement\}

Reference Directory Structure (similar case): \{dir\_structure\}

\{dir\_counts\_str\}

Make sure you generate all the necessary files for the user's requirements.
Please generate the output as structured JSON.
\end{tcolorbox}

\subsection{Input Writer Agent Prompts}

The Input Writer Agent generates \texttt{OpenFOAM} configuration files and ensures consistency across interdependent files.

\subsubsection{File Generation Prompts}

\begin{tcolorbox}[colback=blue!5!white,colframe=blue!75!black,title=File Generation System Prompt]
You are an expert in \texttt{OpenFOAM} simulation and numerical modeling.
Your task is to generate a complete and functional file named: $<$file\_name$>$\{file\_name\}$<$/file\_name$>$ within the $<$folder\_name$>$\{folder\_name\}$<$/folder\_name$>$ directory. 
Ensure all required values are present and match with the files content already generated.
Before finalizing the output, ensure:
- All necessary fields exist (e.g., if `nu` is defined in `constant/transportProperties`, it must be used correctly in `0/U`).
- Cross-check field names between different files to avoid mismatches.
- Ensure units and dimensions are correct** for all physical variables.
- Ensure case solver settings are consistent with the user's requirements. Available solvers are: \{state.case\_stats['case\_solver']\}.
Provide only the code—no explanations, comments, or additional text.
\end{tcolorbox}

\begin{tcolorbox}[colback=blue!5!white,colframe=blue!75!black,title=File Generation User Prompt]
User requirement: \{state.user\_requirement\}
Refer to the following similar case file content to ensure the generated file aligns with the user requirement:
$<$similar\_case\_reference$>$\{similar\_file\_text\}$<$/similar\_case\_reference$>$
Similar case reference is always correct. If you find the user requirement is very consistent with the similar case reference, you should use the similar case reference as the template to generate the file.
Just modify the necessary parts to make the file complete and functional.
Please ensure that the generated file is complete, functional, and logically sound.
Additionally, apply your domain expertise to verify that all numerical values are consistent with the user's requirements, maintaining accuracy and coherence.

[If previously generated files exist:]
The following are files content already generated: \{str(writed\_files)\}

You should ensure that the new file is consistent with the previous files. Such as boundary conditions, mesh settings, etc.
\end{tcolorbox}

\subsubsection{Command Generation Prompts}

\begin{tcolorbox}[colback=blue!5!white,colframe=blue!75!black,title=Command Generation System Prompt]
You are an expert in \texttt{OpenFOAM}. The user will provide a list of available commands. 
Your task is to generate only the necessary \texttt{OpenFOAM} commands required to create an Allrun script for the given user case, based on the provided directory structure. 
Return only the list of commands—no explanations, comments, or additional text.
\end{tcolorbox}

\begin{tcolorbox}[colback=blue!5!white,colframe=blue!75!black,title=Command Generation User Prompt]
Available \texttt{OpenFOAM} commands for the Allrun script: \{commands\}
Case directory structure: \{state.dir\_structure\}
User case information: \{state.case\_info\}
Reference Allrun scripts from similar cases: \{state.allrun\_reference\}
Generate only the required \texttt{OpenFOAM} command list—no extra text.
\end{tcolorbox}

\subsubsection{Allrun Script Generation Prompts}

\begin{tcolorbox}[colback=blue!5!white,colframe=blue!75!black,title=Allrun Script Generation System Prompt]
You are an expert in \texttt{OpenFOAM}. Generate an Allrun script based on the provided details.
Available commands with descriptions: \{commands\_help\}

Reference Allrun scripts from similar cases: \{state.allrun\_reference\}
\end{tcolorbox}

\begin{tcolorbox}[colback=blue!5!white,colframe=blue!75!black,title=Allrun Script Generation User Prompt]
User requirement: \{state.user\_requirement\}
Case directory structure: \{state.dir\_structure\}
User case infomation: \{state.case\_info\}
All run scripts for these similar cases are for reference only and may not be correct, as you might be a different case solver or have a different directory structure. 
You need to rely on your \texttt{OpenFOAM} and physics knowledge to discern this, and pay more attention to user requirements, 
as your ultimate goal is to fulfill the user's requirements and generate an allrun script that meets those requirements.
Generate the Allrun script strictly based on the above information. Do not include explanations, comments, or additional text. Put the code in ``` tags.
\end{tcolorbox}

\subsection{Reviewer Agent Prompts}

The Reviewer Agent analyzes simulation errors and proposes corrections to resolve issues.

\subsubsection{Error Analysis Prompts}

\begin{tcolorbox}[colback=blue!5!white,colframe=blue!75!black,title=Error Analysis System Prompt]
You are an expert in \texttt{OpenFOAM} simulation and numerical modeling. 
Your task is to review the provided error logs and diagnose the underlying issues. 
You will be provided with a similar case reference, which is a list of similar cases that are ordered by similarity. You can use this reference to help you understand the user requirement and the error.
When an error indicates that a specific keyword is undefined (for example, 'div(phi,(p|rho)) is undefined'), your response must propose a solution that simply defines that exact keyword as shown in the error log. 
Do not reinterpret or modify the keyword (e.g., do not treat '|' as 'or'); instead, assume it is meant to be taken literally. 
Propose ideas on how to resolve the errors, but do not modify any files directly. 
Please do not propose solutions that require modifying any parameters declared in the user requirement, try other approaches instead. Do not ask the user any questions.
The user will supply all relevant foam files along with the error logs, and within the logs, you will find both the error content and the corresponding error command indicated by the log file name.
\end{tcolorbox}

\begin{tcolorbox}[colback=blue!5!white,colframe=blue!75!black,title=Error Analysis User Prompt (Initial Error)]
$<$similar\_case\_reference$>$\{state.tutorial\_reference\}$<$/similar\_case\_reference$>$
$<$foamfiles$>$\{str(state.foamfiles)\}$<$/foamfiles$>$
$<$error\_logs$>$\{state.error\_logs\}
$<$/error\_logs$>$
$<$user\_requirement$>$\{state.user\_requirement\}$<$/user\_requirement$>$
Please review the error logs and provide guidance on how to resolve the reported errors. Make sure your suggestions adhere to user requirements and do not contradict it.
\end{tcolorbox}

\begin{tcolorbox}[colback=blue!5!white,colframe=blue!75!black,title=Error Analysis User Prompt (Subsequent Errors)]
$<$similar\_case\_reference$>$\{state.tutorial\_reference\}$<$/similar\_case\_reference$>$
$<$foamfiles$>$\{str(state.foamfiles)\}$<$/foamfiles$>$
$<$current\_error\_logs$>$\{state.error\_logs\}
$<$/current\_error\_logs$>$
$<$history$>$
\{chr(10).join(state.history\_text)\}
$<$/history$>$$<$user\_requirement$>$\{state.user\_requirement\}$<$/user\_requirement$>$

I have modified the files according to your previous suggestions. If the error persists, please provide further guidance. Make sure your suggestions adhere to user requirements and do not contradict it. Also, please consider the previous attempts and try a different approach.
\end{tcolorbox}

\subsubsection{File Correction Prompts}

\begin{tcolorbox}[colback=blue!5!white,colframe=blue!75!black,title=File Correction System Prompt]
You are an expert in \texttt{OpenFOAM} simulation and numerical modeling. 
Your task is to modify and rewrite the necessary \texttt{OpenFOAM} files to fix the reported error. 
Please do not propose solutions that require modifying any parameters declared in the user requirement, try other approaches instead.
The user will provide the error content, error command, reviewer's suggestions, and all relevant foam files. 
Only return files that require rewriting, modification, or addition; do not include files that remain unchanged. 
Return the complete, corrected file contents in the following JSON format: 
list of foamfile: [\{file\_name: 'file\_name', folder\_name: 'folder\_name', content: 'content'\}]. 
Ensure your response includes only the modified file content with no extra text, as it will be parsed using Pydantic.
\end{tcolorbox}

\begin{tcolorbox}[colback=blue!5!white,colframe=blue!75!black,title=File Correction User Prompt]
$<$foamfiles$>$\{str(state.foamfiles)\}$<$/foamfiles$>$
$<$error\_logs$>$\{state.error\_logs\}$<$/error\_logs$>$
$<$reviewer\_analysis$>$\{review\_content\}$<$/reviewer\_analysis$>$

$<$user\_requirement$>$\{state.user\_requirement\}$<$/user\_requirement$>$

Please update the relevant \texttt{OpenFOAM} files to resolve the reported errors, ensuring that all modifications strictly adhere to the specified formats. Ensure all modifications adhere to user requirement.
\end{tcolorbox}

\subsection{History Tracking Format}

The system tracks modification history using a structured format for each iteration attempt:
\begin{tcolorbox}[colback=blue!5!white,colframe=blue!75!black,title=History Tracking Format]
$<$Attempt \{attempt\_number\}$>$
$<$Error\_Logs$>$
\{state.error\_logs\}
$<$/Error\_Logs$>$
$<$Review\_Analysis$>$
\{review\_content\}
$<$/Review\_Analysis$>$
$<$/Attempt$>$
\end{tcolorbox}

\subsection{Example User Requirements}

Below is an example of a user requirement used to test the Foam-Agent system:

\begin{tcolorbox}[colback=blue!5!white,colframe=blue!75!black,title=Example User Requirement]
Perform a 3D Bernard Cell simulation using \texttt{OpenFOAM}. 
The computational domain spans 9 m x 1 m x 2 m. 
The simulation begins at t=0 seconds and runs until t=1000 seconds with a time step of 1 second, 
and results are written at intervals of every 50 seconds. One wall has a temperature of 301 K, while the other has a temperature of 300 K.
\end{tcolorbox}

\section{User Prompts In Case Studies}
\subsection{Multi Element Airfoil}\label{prompt:multi_airfoil}
\begin{tcolorbox}[colback=blue!5!white,colframe=blue!75!black,title=Example User Requirement]
do an incompressible 2D incompressible flow over a multi element airfoil setup. The mesh is provided as a .msh file. The msh file contains 4 boundaries named "inlet",
"outlet", "walls", "airfoil" and "frontAndBack". The "inlet" and "outlet" are of type freestream with the freestream velocity being 9 m/s. 
 The "walls" and "airfoil" have a no-slip boundary condition
  (velocity equal to zero at the wall). The "frontAndBack"  faces are designated as
  'empty'. The simulation runs from time 0 to 10 with a time step of 1.0 units,
  and results are output every 1 time steps. The viscosity (nu) is set as constant
  with a value of $1.5\times10^{-5}$ $\mathrm{m^2/s}$. Use simpleFoam solver. Use SpalartAllmaras turbulence model. Further visualize the magnitude of velocity along the Z plane.
\end{tcolorbox}

\subsection{Tandem Wing}\label{prompt:tandem_wing}
\begin{tcolorbox}[colback=blue!5!white,colframe=blue!75!black,title=Example User Requirement]
do an incompressible 3D incompressible flow over a tandem wing configuration. The mesh is provided as a .msh file. The msh file contains 4 boundaries named "inlet",
"outlet", "walls", "airfoil" and "frontAndBack". The "inlet" and "outlet" are of type freestream with the freestream velocity being 9 m/s. 
 The "walls" and "airfoil" have a no-slip boundary condition (velocity equal to zero at the wall). The "frontAndBack"  faces are also of type wall. 
 The simulation runs from time 0 to 10 with a time step of 1.0 units,
  and results are output every 1 time steps. The viscosity (`nu`) is set as constant
  with a value of $1.5\times10^{-5}$ $\mathrm{m^2/s}$. Use simpleFoam solver. Use SpalartAllmaras turbulence model. Further visualize the magnitude of velocity along the mid Z section at the final time.
\end{tcolorbox}

\subsection{Flow Over Cylinder}\label{prompt:cylinder_flow}
\begin{tcolorbox}[colback=blue!5!white,colframe=blue!75!black,title=Example User Requirement]
Simulate incompressible flow over a circular cylinder. 
Use gmsh to create the computational mesh.
The computational domain extends from -2.5 to 2.5 in the x-direction, -1 to 1 in the y-direction, and 0 to 0.2 in the z-direction. 
The cylinder is positioned at (-1, 0) with a radius of 0.1 units. 
Use a structured mesh with approximately 20x10 cells in the x-y plane and 1 cell in the z-direction. 
The inlet boundary named "inlet" (left boundary at x = -2.5) has a uniform velocity of 1 m/s in the positive x-direction. 
The right boundary at x=+2.5 is the outlet named "outlet". 
The top and bottom walls named "topWall" and "bottomWall" respectively (y = +1 and y=-1) use slip boundary conditions. 
The cylinder surface named "cylinder" uses a no-slip boundary condition (velocity equal to zero at the wall). 
The front and back faces named "frontAndBack" are located at z = 0 and z = 0.2 respectively, and are designated as 'empty' for 2D simulation.  
Use base mesh size of 0.5 on cylinder and size of 1.0 elsewhere.
The simulation runs from time 0 to 2 seconds with a time step of 0.001 units, and results are output every 100 time steps. 
The kinematic viscosity (nu) is set as constant with a value of $1\times10^{-5}$ $\mathrm{m^2/s}$. Use pisoFoam solver for incompressible flow.
Visualize the magnitude of velocity ('U') along the x-y plane. 
\end{tcolorbox}

\subsection{Flow Over Two Square Obstacles}\label{prompt:two_squares}
\begin{tcolorbox}[colback=blue!5!white,colframe=blue!75!black,title=Example User Requirement]
Simulate incompressible flow over two square obstacles. 
Use gmsh to create the computational mesh.
The computational domain spans 0 to 5 in x direction and 0 to 2.5 in y direction and 0 to 0.1 in z direction. 
One of the square obstacle is of size 0.25 unit x 0.25 unit x 0.1 unit centered at 1.5 x 1.25 x 0.0 and the other square obstacle is of size 0.25 unit x 0.25 unit x 0.1 unit centered at 3.5 x 1.25 x 0.0. 
Use one cell in z direction making the geometry effectively 2D. 
Use a structured mesh with approximately 50x25 cells in the x-y plane and 1 cell in the z-direction. 
The inlet boundary named "inlet" (left boundary at x = 0) has a uniform velocity of 1 m/s in the positive x-direction. 
The right boundary at x = 5 is the outlet named "outlet". 
The top and bottom walls named "topWall" and "bottomWall" respectively (y = 2.5 and y = 0) use slip boundary conditions. 
The square obstacle surfaces named "square1" and "square2" use no-slip boundary conditions (velocity equal to zero at the walls). 
The front and back faces named "frontAndBack" are located at z = 0 and z = 0.1 respectively, and are designated as 'empty' for 2D simulation.  
Use base mesh size of 0.5 on squares and size of 1.0 elsewhere.
The simulation runs from time 0 to 10 seconds with a time step of 0.001 units, and results are output every 100 time steps. 
The kinematic viscosity (nu) is set as constant with a value of $1\times10^{-5}$ $\mathrm{m^2/s}$. Use pisoFoam solver for incompressible flow.
Visualize the magnitude of velocity ('U') along the x-y plane. 
\end{tcolorbox}

\subsection{3D cavity HPC Case}\label{prompt:hpc_cavity}
\begin{tcolorbox}[colback=blue!5!white,colframe=blue!75!black,title=Example User Requirement]
Do an incompressible 3D lid driven cavity flow using icoFoam solver. 
The cavity is a cube of dimension [0, 0.1]x [0, 0.1]x [0,0.1]. 
Use simple grading with 100X100x100 in x, y and z direction. 
The top wall ('movingWall') moves in the x-direction with a uniform velocity of 1 m/s. 
The 'fixedWalls' have a no-slip boundary condition (velocity equal to zero at the wall). 
The simulation runs from time 0 to 0.015 with a time step of 0.001 units, and results are output every 10 time steps. 
The viscosity (nu) is set as constant with a value of 0.01 $\mathrm{m^2/s}$.
Perform an hpc run for this case in perlmutter cluster. My account is xxxx. Do a parallel run for this case by splitting it into 32 subdomains. 
\end{tcolorbox}

\section{Slurm Script Generated by the Agent}        
        
        \begin{tcolorbox}[colback=blue!5!white,colframe=blue!75!black,title=Slurm Script Generated by the Agent]

\#!/bin/bash\\
\#SBATCH -A xxxx\\
\#SBATCH -C cpu\\
\#SBATCH -q regular\\
\#SBATCH -N 1\\
\#SBATCH -n 32\\
\#SBATCH --ntasks-per-node=32\\
\#SBATCH -t 02:00:00\\
\#SBATCH --mem=128GB\\
\#SBATCH -J Sim\\
\#SBATCH -o \%j.out\\
\#SBATCH -e \%j.err\\

\# Initialize error handling\\
set -e\\

\# Change to case directory\\
cd ./runshpcnode || exit 1\\

\# Create log directory if it doesn't exist\\
mkdir -p logs\\

\# Run the simulation\\
echo "Starting \texttt{OpenFOAM} simulation at \$(date)"\\
./Allrun -parallel\\
if [ \$? -eq 0 ]; then\\
    echo "Simulation completed successfully at \$(date)"\\
    exit 0\\
else\\
    echo "Simulation failed at \$(date)"\\
    exit 1\\
fi
        \end{tcolorbox}\label{slurm_script}

\end{appendices}

\end{document}